# Can AI help in screening Viral and COVID-19 pneumonia?

Muhammad E. H. Chowdhury[1]*, Tawsifur Rahman[2], Amith Khandakar[1], Rashid Mazhar[3], Muhammad Abdul Kadir[2], Zaid Bin Mahbub[4], Khandakar R. Islam[5], Muhammad Salman Khan[6,7], Atif Iqbal[1], Nasser Al-Emadi[1], Mamun Bin Ibne Reaz[8], M. T. Islam[8]

[1]Department of Electrical Engineering, Qatar University, Doha-2713, Qatar
[2]Department of Biomedical Physics & Technology, University of Dhaka, Dhaka-1000, Bangladesh
[3]Thoracic Surgery, Hamad General Hospital, Doha-3050, Qatar
[4]Department of Mathematics and Physics, North South University, Dhaka-1229, Bangladesh
[5]Department of Orthodontics, Bangabandhu Sheikh Mujib Medical University, Dhaka-1000, Bangladesh
[6]Department of Electrical Engineering (JC), University of Engineering and Technology, Peshawar-25120, Pakistan
[7]Artificial Intelligence in Healthcare, Intelligent Information Processing Lab, National Center for Artificial Intelligence, University of Engineering and Technology, Peshawar, Pakistan
[8]Department of Electrical, Electronic & Systems Engineering, Universiti Kebangsaan Malaysia, Bangi, Selangor 43600, Malaysia

*Correspondence: Dr. Muhammad E. H. Chowdhury; mchowdhury@qu.edu.qa, Tel.: +974-31010775

**ABSTRACT:** Coronavirus disease (COVID-19) is a pandemic disease, which has already caused thousands of causalities and infected several millions of people worldwide. Any technological tool enabling rapid screening of the COVID-19 infection with high accuracy can be crucially helpful to the healthcare professionals. The main clinical tool currently in use for the diagnosis of COVID-19 is the Reverse transcription polymerase chain reaction (RT-PCR), which is expensive, less-sensitive and requires specialized medical personnel. X-ray imaging is an easily accessible tool that can be an excellent alternative in the COVID-19 diagnosis. This research was taken to investigate the utility of artificial intelligence (AI) in the rapid and accurate detection of COVID-19 from chest X-ray images. The aim of this paper is to propose a robust technique for automatic detection of COVID-19 pneumonia from digital chest X-ray images applying pre-trained deep-learning algorithms while maximizing the detection accuracy. A public database was created by the authors combining several public databases and also by collecting images from recently published articles. The database contains a mixture of 423 COVID-19, 1485 viral pneumonia, and 1579 normal chest X-ray images. Transfer learning technique was used with the help of image augmentation to train and validate several pre-trained deep Convolutional Neural Networks (CNNs). The networks were trained to classify two different schemes: i) normal and COVID-19 pneumonia; ii) normal, viral and COVID-19 pneumonia with and without image augmentation. The classification accuracy, precision, sensitivity, and specificity for both the schemes were 99.7%, 99.7%, 99.7% and 99.55% and 97.9%, 97.95%, 97.9%, and 98.8%, respectively. The high accuracy of this computer-aided diagnostic tool can significantly improve the speed and accuracy of COVID-19 diagnosis. This would be extremely useful in this pandemic where disease burden and need for preventive measures are at odds with available resources.

## I. INTRODUCTION

Coronavirus disease (COVID-19) is an extremely contagious disease and it has been declared as a pandemic by the World Health Organization (WHO) on 11th March 2020 considering the extent of its spread throughout the world [1]. The pandemic declaration also stressed the deep concerns of the alarming rate of spread and severity of COVID-19. It is the first recorded pandemic caused by any coronavirus. It is defined as a global health crisis of its time, which has spread all over the world. Governments of different countries have imposed border restrictions, flight restrictions, social distancing, and increasing awareness of hygiene. However, the virus is still spreading at very rapid rate. While most of the people infected with the COVID-19 experienced mild to moderate respiratory illness, some developed a deadly pneumonia. There are assumptions that elderly people with underlying medical problems like cardiovascular disease, diabetes, chronic respiratory disease, renal or hepatic diseases and cancer are more likely to develop serious illness [2]. Until now, no specific vaccine or treatment for COVID-19 has been invented. However, there are many ongoing clinical trials evaluating potential treatments. More than 7.5 million infected cases were found in more than 200 countries until 11th June 2020, among which around 421 thousand deaths, 3.8 million recovery, 3.2 million mild cases and 54 thousand critical cases were reported [3, 4].

In order to combat with the spreading of COVID-19, effective screening and immediate medical response for the infected patients is a crying need. Reverse Transcription Polymerase chain reaction (RT-PCR) is the most used clinical screening method for the COVID-19 patients, which uses respiratory specimens for testing [5]. RT-PCR is used as a reference method for the detection of COVID-19 patients, however, the technique is manual, complicated, laborious and time-consuming with a positivity rate of only 63% [5]. Moreover, there is a significant shortage of its supply, which leads to delay in the disease prevention efforts [6]. Many countries are facing difficulties with incorrect number of COVID-19 positive cases because of not only due to the lack of test kits but also due to the delay in the test results [7]. These delays can lead to infected patients interacting with the healthy patients and infecting them in the process. It is reported that the RT-PCR kit costs about USD 120-130 and also requires a specialized biosafety lab to house the PCR machine, each of which may cost USD 15,000 to USD 90,000 [8]. Such an expensive screening tool with delayed test results is leading to spread of the disease, making the scenario worst. This is not an issue for the low-income countries only but certain developed countries are also struggling to tackle with this [9]. The other diagnosis methods of the COVID-19 include clinical symptoms analysis, epidemiological history and positive radiographic images (computed tomography (CT) /Chest radiograph (CXR)) as well as positive pathogenic testing. The clinical characteristics of severe COVID-19 infection is that of bronchopneumonia causing fever, cough, dyspnea, and respiratory failure with acute respiratory distress syndrome (ARDS) [10-13]. Readily available radiological imaging is an important diagnostic tool for COVID-19. The majority of COVID-19 cases have similar features on radiographic images including bilateral, multi-focal, ground-glass opacities with a peripheral or posterior distribution, mainly in the lower lobes, in the early stage and pulmonary consolidation in the late stage [13-19]. Although typical CXR images may help early screening of suspected cases, the images of various viral pneumonias are similar and they overlap with other infectious and inflammatory lung diseases. Therefore, it is difficult for radiologists to distinguish COVID-19 from other viral pneumonias. The symptoms of COVID-19 being similar to that of viral pneumonia can sometimes lead to wrong diagnosis in the current situation while hospitals are overloaded and working round the clock. Such an incorrect diagnosis can lead to a non-COVID viral Pneumonia being falsely labelled as highly suspicious of having COVID-19 and thus delaying in treatment with consequent costs, effort and risk of exposure to positive COVID-19 patients.

Currently many biomedical health problems and complications (e.g. brain tumor detection, breast cancer detection, etc.) are using Artificial Intelligence (AI) based solutions [20-25]. Deep learning techniques can reveal image features, which are not apparent in the original images. Specifically, Convolutional Neural Network (CNN) has been proven extremely beneficial in feature extraction and learning and therefore, widely adopted by the research community [26]. CNN was used to enhance image quality in low-light images from a high-speed video endoscopy [27] and was also applied to identify the nature of pulmonary nodules via CT images, the diagnosis of pediatric pneumonia via chest X-ray images, automated labelling of polyps during colonoscopic videos, cystoscopic image analysis from videos [28-31]. Deep learning techniques on chest X-Rays are getting popularity with the availability of the deep CNNs and the promising results it has shown in different applications. Moreover, there is an abundance of data available for training different machine-learning models. Transfer learning technique has significantly eased the process by allowing quickly retrain a very deep CNN network with a comparatively low number of images. Concept of transfer learning in deep learning framework was used by Vikash et al.[32] for the detection of pneumonia using pre-trained ImageNet models [33] and their ensembles. A customized VGG16 model was used by Xianghong et al. [34] for lung regions identification and different types of pneumonia classification. Wang et al.[35] used a large hospital-scale dataset for classification and localization of common thoracic diseases and Ronneburger et al.[36] used image augmentation on a small set of images to train deep CNN for image segmentation problem to achieve better performance. Rajpurkar et al.[37] reported a 121-layer CNN (CheXNet) on chest X-rays to detect 14 different pathologies, including pneumonia using an ensemble of different networks. A pre-trained DenseNet-121 and feature extraction techniques were used in the accurate identification of 14 thoracic diseases in [38]. Sundaram et al. [39] used AlexNet and GoogLeNet with image augmentation to obtain an Area Under the Curve (AUC) of 0.95 in pneumonia detection.

Recently, several groups have reported deep machine learning techniques using X-ray images for detecting COVID-19 pneumonia [40-57]. However, most of these groups used rather a small dataset containing only a few COVID-19 samples. This makes it difficult to generalize their results reported in these articles and cannot guarantee that the reported performance will retain when these models will be tested on a larger dataset. Ioannis *et al.* [39] reported transfer learning approach for classifying dataset of 1427 X-ray images containing 224 COVID-19, 700 Bacterial Pneumonia and 504 Normal X-ray images with accuracy, sensitivity, and specificity of 96.78%, 98.66%, and 96.46% respectively. Different pre-trained models were compared however, the reported results were based on a small dataset. Ashfar *et al.* [43] proposed a Capsule Networks, called COVID-CAPS rather than a conventional CNN to deal with a smaller dataset. COVID-CAPS was reported to achieve an accuracy of 95.7%, sensitivity of 90%, and specificity of 95.8%. Abbas *et al.* [44] have worked on a very small database of 105 COVID-19, 80 Normal and 11 SARS X-ray images to detect COVID-19 X-ray images using modified pre-trained CNN model (*DeTraC-Decompose, Transfer and Compose*) to project the high-dimension feature space into a lower one. This would help to produce more homogenous classes, lessen the memory requirements and achieved accuracy, sensitivity and specificity of 95.12%, 97.91% and 91.87% respectively. Wang and Wong in [40] introduced a deep CNN, called COVID-Net for the detection of COVID-19 cases from around 14k chest X-ray images, however the achieved accuracy was 83.5%. Ucar *et al.* [47] has fine-tuned SqueezeNet pre-trained network with Bayesian optimization to classify COVID-19 images, which showed promising result on a small dataset. This approach should be evaluated on a large COVID and non-COVID dataset. Khan *et al.* [52] applied transfer learning approach on 310 normal, 330 bacterial pneumonia, 327 viral pneumonia and 284 COVID-19 pneumonia images. However, different machine learning algorithms were not evaluated in this study and the experimental protocol was not clear in this work.

In summary, several recent works were reported on transfer learning approach for the detection of COVID-19 X-ray images from a small dataset with promising results however these needed to be verified on a large dataset. Some group have modified or fine-tuned the pre-trained networks to achieve better performance while some groups use capsule networks. A rigorous experiment on a large database of COVID and non-COVID classes are very few and missing in case of transfer learning approach. The authors in this paper have prepared a large database of X-ray images of 1579 normal, 1485 viral pneumonia and 423 COVID-19 positive pneumonia and made this publicly available so that other researchers can get benefit from it. Moreover, eight different pre-trained deep learning networks were trained, validated and tested for two different classification schemes. One classification model was trained to classify COVID-19 and normal X-ray images while other was trained to classify normal, viral pneumonia and COVID-19 pneumonia images. Both of the experiments were evaluated with and without image augmentation technique to study the effect of image augmentation in this particular problem.

## II. METHODOLOGY

Deep convolutional neural networks typically perform better with a larger dataset than a smaller one. Transfer learning can be used in the training of deep CNNs where the dataset is not large. The concept of transfer learning uses the trained model from large dataset such as ImageNet [58] and modify the Softmax and classification layer of the pre-trained networks. The pre-trained weights are then used for faster training of the network for an application with comparatively smaller dataset. This removes the requirement of having large dataset and also reduces the long training period as is required by the deep learning algorithm when developed from scratch [59, 60].

Although there are a large number of COVID-19 patients infected worldwide, the number of chest X-ray images publicly available online are small and scattered. Therefore, in this work, authors have reported a comparatively large dataset of COVID-19 positive chest X-ray images while normal and viral pneumonia images are readily available publicly and used for this study. A Kaggle database was created by the authors to make the database publicly available to the researchers worldwide and the trained models were made available so that others can get benefit of this study [61].

### A. DATABASE DESCRIPTION

In this study, posterior-to-anterior (AP)/anterior-to-posterior (PA) image of chest X-ray was used as this view of radiography is widely used by radiologist in clinical diagnosis. Six different sub-databases were used to create one database. Among these databases, COVID-19 database was developed by the authors from collected and publicly available databases, while normal and viral pneumonia databases were created from publicly available Kaggle databases. In the following section, authors have summarized how this dataset is created.

COVID-19 sub-database, comprising of 423 AP/PA images, was created from the following four major data sources.

- *Italian Society of Medical and Interventional Radiology (SIRM) COVID-19 DATABASE:*

SIRM COVID-19 database [62] reports 384 COVID-19 positive radiographic images (CXR and CT) with varying resolution. Out of 384 radiographic images, 94 images are chest X-ray images and 290 images are lung CT images. This database is updated in a random manner and until 10th May 2020, there were 71 confirmed COVID-19 cases were reported in this database.

- *Novel Corona Virus 2019 Dataset:*

Joseph Paul Cohen and Paul Morrison and Lan Dao have created a public database in GitHub [63] by collecting 319 radiographic images of COVID-19, Middle East respiratory syndrome (MERS), Severe acute respiratory syndrome (SARS) and ARDS from the published articles and online resources. In this database, they have collected 250 COVID-19 positive chest X-ray images and 25 COVID-19 positive lung CT images with varying image resolutions. However, in this study, authors have considered 134 COVID-19 positive chest X-ray images, which are different from the images of the database that the authors created from different articles.

- *COVID-19 positive chest x-ray images from different articles:*

GitHub database has encouraged the authors to look into the literature and interestingly more than 1200 articles were published in less than two-months of period. Authors have observed that the GitHub database has not collected most of the X-ray and CT images rather a small number of images were in that database. Moreover, the images in SIRM and GitHub database are in random size depending on the X-ray machine resolution and the articles from which it was taken. Therefore, authors have carried out a tedious task of collecting and indexing the X-ray and CT images from all the recently publicly available articles and online sources. These articles and the radiographic images were then compared with the GitHub database to avoid duplication. Authors managed to collect 60 COVID-19 positive chest X-ray images from 43 recently published articles [61], which were not listed in the GitHub database and 32 positive chest x-ray images from Radiopaedia [64], which were not listed in the GitHub database.

- *COVID-19 Chest imaging at thread reader*

A physician has shared 103 images for 50 different cases with varying resolution from his hospital in Spain to the Chest imaging at thread reader [65].Images from RSNA-Pneumonia-Detection-Challenge database along with the Chest X-ray Images database from Kaggle were used to create the normal and viral pneumonia sub-databases of 1579 and 1485 X-ray images respectively.

- *RSNA-Pneumonia-Detection-Challenge*

In 2018, Radiology Society of North America (RSNA) organized an artificial intelligence (AI) challenge to detect pneumonia from the chest X-ray images. In this database, normal chest X-ray with no lung infection and non-COVID pneumonia images were available [66].

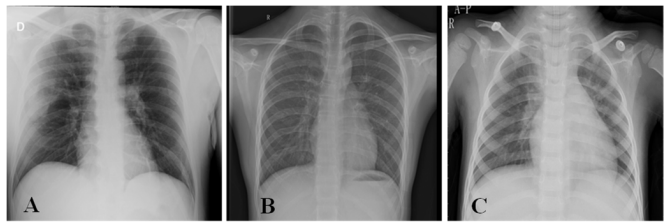

**Figure 1:** Sample X-ray image from the dataset: COVID-19 X-ray image (A), normal X-ray image (B), and Viral Pneumonia X-ray image (C).

- *Chest X-Ray Images (pneumonia):*

Kaggle chest X-ray database is a very popular database, which has 5247 chest X-ray images of normal, viral and bacterial pneumonia with resolution varying from 400p to 2000p [67]. Out of 5247 chest X-ray images, 3906 images are from different subjects affected by pneumonia (2561 images for bacterial pneumonia and 1345 images for viral pneumonia) and 1341 images are from normal subjects. Chest X-ray images for normal and viral pneumonia were used from this database to create the new database. Figure 1 shows sample images from the database for normal, COVID-19 pneumonia, and viral pneumonia chest X-ray images.

### B. CNN Model Selection

Eight different pre-trained CNN models were trained, validated and tested in this study. The experimental evaluation of MobileNetv2, SqueezeNet[68], ResNet18[69], ResNet101 and DenseNet201 were performed utilizing MATLAB 2020a running on a computer with Intel© i7-core @3.6GHz processor and 16GB RAM, with an 8-GB NVIDIA GeForce GTX 1080 graphics processing unit (GPU) card on 64-bit Windows 10 operating system. On the other hand, CheXNet, Inceptionv3 and VGG19 were implemented using PyTorch library with Python on Intel® Xeon® CPU E5-2697 v4 @ 2,30GHz and 64 GB RAM, with a 16 GB NVIDIA GeForce GTX 1080 GPU. Three comparatively shallow networks (MobileNetv2, SqueezeNet and ResNet18) and five deep networks (Inceptionv3, ResNet101, CheXNet, VGG19 and DenseNet201) were evaluated in this study to investigate whether shallow or deep networks are suitable for this application. Two different variants of ResNet were used to compare specifically the impact of shallow and deep networks with similar structure. Performance difference due to initially trained on different image classes other than X-ray images were compared with CheXNet, which is a 121-layer DenseNet variant and the only network pre-trained on X-ray images. Several researchers showed the reliability of using this network for COVID-19 classification. Therefore, it was important to investigate whether CheXNet outperforms other deep networks or not. Eight pre-trained CNN models were trained using stochastic Gradient Descent (SGD) with momentum optimizer with learning rate, $\alpha = 10^{-3}$, momentum update, $\beta = 0.9$ and mini-batch size of 16 images with 20 Back Propagation epochs. Fivefold cross-validation result was averaged to produce the final receiver operating characteristic (ROC) curve, confusion matrix, and evaluation matrices.

Two different experiments were carried out in this study: i) Two-class image classification using models trained without and with images augmentation, and ii) Three-class image classification using models trained without and with image augmentation. Figure 2 illustrates the overall system diagram with the three-class image classification problem.

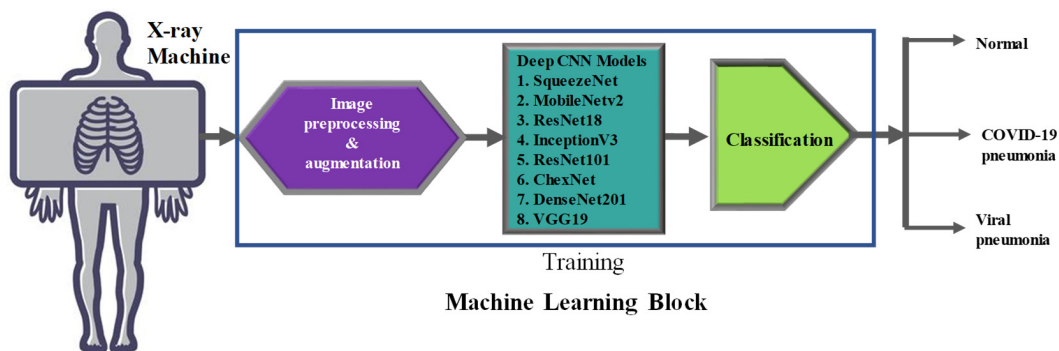

**Figure 2**: Block diagram of the overall system.

TABLE 1
NUMBER OF IMAGES PER CLASS AND PER FOLD BEFORE AND AFTER DATA AUGMENTATION.

| Types | Total No. of X-ray images/class | Training without augmentation | | | | Training with image augmentation | | | |
|---|---|---|---|---|---|---|---|---|---|
| | | Used Image | Training set/fold | Validation/fold | Test set | Training set/fold | Augmented image/fold | Validation/fold | Test image/fold |
| COVID-19 | 423 | 423 | 304 | 34 | 85 | 304 | 2128 | 34 | 85 |
| Normal | 1579 | 423 | 304 | 34 | 85 | 1137 | 2274 | 126 | 316 |
| Viral Pneumonia | 1485 | 423 | 304 | 34 | 85 | 1069 | 2138 | 119 | 297 |

### B. PREPROCESSING

Chest X-ray images were only resized before applying as input to the networks. Input requirements for different CNNs are different. For SqueezeNet, the images were resized to 227×227 pixels whereas for mobilenetv2, ResNet18, ResNet101, VGG19 and DenseNet201, the images were resized to 224×224 pixels; and for Inceptionv3 the images were resized to 299×299 pixels. All images were normalized according to the pre-trained model standards.

In the study1, image augmentation technique was not applied to the training data. Since COVID-19 positive chest X-ray images were 423, same number of X-ray images were randomly selected from normal (out of 1579) and viral pneumonia (out of 1485) images to match with COVID-19 images to balance the database. In study2, entire database (i.e., 423 COVID-19, 1579 normal and 1485 viral pneumonia images) was used. Both the experiments were evaluated using a stratified 5-fold cross-validation (CV) scheme with a ratio of 80% for training and 20% for the test (unseen folds) splits, where 10% of training data is used as a validation set to avoid overfitting. However, in study2, COVID-19 images are much smaller in number than that in the other two image classes. Moreover, overall image number in any class was not several thousand. Therefore, Image augmentation techniques were applied to viral pneumonia, normal and COVID-19 X-ray images for training to create a balanced training set. However, COVID-19 images were augmented six times while normal and viral pneumonia images were augmented once only.

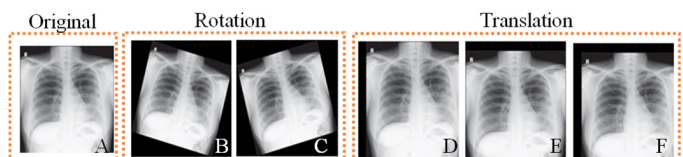

**Figure 3**: Original Chest X-ray image (A), Image after rotation by 15 degree clockwise (B), Image after rotation by 15 degree counter clockwise (C), Image after 5% horizontal translation (D), after 5% vertical translation (E), and after 5% horizontal and vertical translation (F).

### C. IMAGE AUGMENTATION

In this study, two different image augmentation techniques (rotation, and translation) were utilized to generate COVID-19 training images, as shown in Figure 3. The rotation operation used for image augmentation was done by rotating the images in the clockwise and counter clockwise direction with an angle of 5, 10 and 15 degrees. Image translation was done by translating image horizontally and vertically by -5% to 5%. However, only image translation was applied to the viral and normal X-ray training images. Table 1 summarizes the number of images per class used for training, validation, and testing at each fold. Study1 was carried out with COVID-19 and normal images while study2 was carried out with COVID-19, normal and viral pneumonia images.

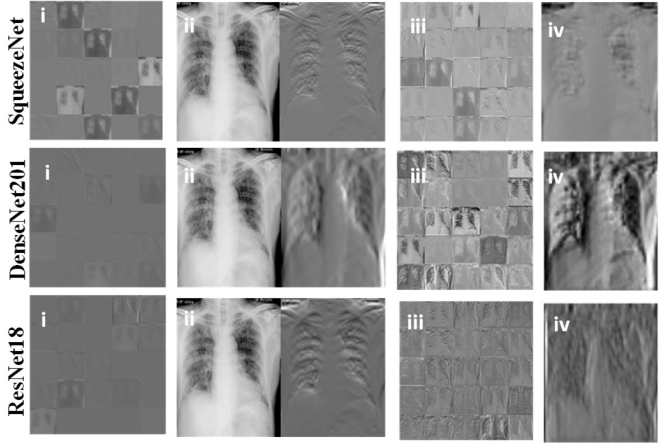

**Figure 4:** Activation map for sample network models of (i) first convolutional layer, (ii) strongest activation channel of first convolutional layer, (iii) deep layer images set, and (iv) corresponding strongest activation channel for the deep convolutional layer for a specific X-ray image input.

### D. INVESTIGATION OF THE DEEP LAYER FEATURES

The deep layers features of the image were investigated by comparing the activated areas of the convolutional layers with the matching regions in the original images. The activation map can take different range of values and was therefore normalized between 0 and 1. The strongest activation channels from the COVID-19, normal and viral pneumonia X-ray images were identified and compared with the original images. It was noticed that the strongest channel activates on edges with positive activation on light left/dark right edges, and negative activation on dark left/light right edges.

Convolutional neural networks learn to detect features like color and edges in their first convolutional layer. In deeper convolutional layers, the network learns to detect features that are more complicated. Later layers build up their features by combining features of earlier layers. Figure 4 shows the activation map in early convolutional layers, deep convolutional layer and their corresponding strongest activation channel for each of the models. It might be difficult to distinguish COVID-19 and viral pneumonia from the original images as reported by different research groups. However, the deep layer features explain better the reason of a deep learning network's failure or success in a particular decision. It provides a visual explanation of the prediction of CNN and it highlights the regions of the images which are contributing more in classification. This technique will be used in the result and discussion section to illustrate how this activation mapping is a distinguishing feature of COVID-19 X-ray images from the other two class of images.

### E. PERFORMANCE EVALUATION MATRIX

In order to evaluate the performance of different deep learning algorithms for classifying the X-ray images in case of two different classification schemes. The trained algorithms were validated using 5-fold cross-validation. The performance of different networks was evaluated using five performances metrics such as- accuracy, sensitivity or recall, specificity, precision (PPV), and F1 score. Per-class values were computed over the overall confusion matrix that accumulates all test fold results of the 5-fold cross-validation.

$$Accuracy_{class\_i} = \frac{TP_{class\_i} + TN_{class\_i}}{TP_{class\_i} + TN_{class\_i} + FP_{class\_i} + FN_{class\_i}} \quad (1)$$

$$Precision_{class\_i} = \frac{TP_{class\_i}}{TP_{class\_i} + FP_{class\_i}} \quad (2)$$

$$Sensitivity_{class_i} = \frac{TP_{class_i}}{TP_{class_i} + FN_{class_i}} \quad (3)$$

$$F1\_score_{class_i} = 2\frac{Precision_{class_i} \times Sensitivity_{class_i}}{Precision_{class_i} + Sensitivity_{class_i}} \quad (4)$$

$$Specificity_{class\_i} = \frac{TN_{class\_i}}{TN_{class\_i} + FP_{class\_i}} \quad (5)$$

where $class_i = COVID-19, and\ Normal$ for two class problem; $COVID-19, Normal\ and\ Viral\ Pneumonia$ for three class problem.

### III. RESULTS AND DISCUSSION

Two different schemes were studied in this study. Classification of COVID-19 and Normal images using eight different pre-trained CNN models while training was done with and without image augmentation. COVID-19, normal and viral pneumonia images were classified using same eight pre-trained models and training was carried out with and without image augmentation.

### A. EXPERIMENTAL RESULTS – TWO CLASS PROBLEM

The comparative performance for different CNNs for two-class classification problem with and without augmentation is shown in Table 2 and comparative AUC curves are shown in Figure 5. It is apparent from Table 2 that all the evaluated pre-trained models perform very well in classifying COVID-19 and normal images in two-class problem. The weighted average performance matrix for eight different networks are very similar whereas small gain can be observed when training was done using image augmentation. Among the networks trained with 338 X-ray images for two-class problem, ResNet18 and CheXNet are equally performing for classifying images while CheXNet and DenseNet201 are performing better than others in case of training with augmented images, although the difference is marginal. CheXNet is producing the highest accuracy of 99.4% and 99.7% for two-class classification without and with image augmentation respectively. Interestingly, CheXNet is performing well in both the cases, with and without augmentation and this can be explained from the fact that CheXNet is the only network which is pre-trained on a large

TABLE 2
WEIGHTED AVERAGE PERFORMANCE METRICS FOR DIFFERENT DEEP LEARNING NETWORKS FOR TWO-CLASS CLASSIFICATION PROBLEM WITH AND WITHOUT IMAGE AUGMENTATION.

| Schemes | Models | Accuracy | Precision (PPV) | Sensitivity (Recall) | F1 Scores | Specificity |
|---|---|---|---|---|---|---|
| Without image augmentation | SqueezeNet | 99.29 | 99.3 | 99.29 | 99.29 | 99.29 |
| | MobileNetv2 | 99.4 | 99.41 | 99.4 | 99.41 | 99.4 |
| | **ResNet18** | **99.41** | **99.42** | **99.41** | **99.41** | **99.41** |
| | InceptionV3 | 99.41 | 100 | 98.81 | 99.4 | 100 |
| | ResNet101 | 99.05 | 99.08 | 99.05 | 99.07 | 99.05 |
| | **CheXNet** | **99.41** | **99.42** | **99.41** | **99.41** | **99.41** |
| | DenseNet201 | 99.3 | 99.4 | 97 | 97.8 | 99.75 |
| | VGG19 | 99.41 | 99.76 | 99.05 | 99.4 | 99.76 |
| With image augmentation | SqueezeNet | 99.40 | 99.40 | 99.40 | 99.40 | 98.84 |
| | MobileNetv2 | 99.65 | 99.65 | 99.65 | 99.65 | 99.26 |
| | ResNet18 | 99.60 | 99.60 | 99.60 | 99.60 | 99.31 |
| | InceptionV3 | 99.40 | 98.80 | 98.33 | 98.56 | 99.70 |
| | ResNet101 | 99.60 | 99.60 | 99.60 | 99.60 | 99.31 |
| | **CheXNet** | **99.69** | **99.69** | **99.69** | **99.69** | **99.23** |
| | **DenseNet201** | **99.70** | **99.70** | **99.70** | **99.70** | **99.55** |
| | VGG19 | 99.60 | 99.20 | 98.60 | 98.90 | 99.80 |

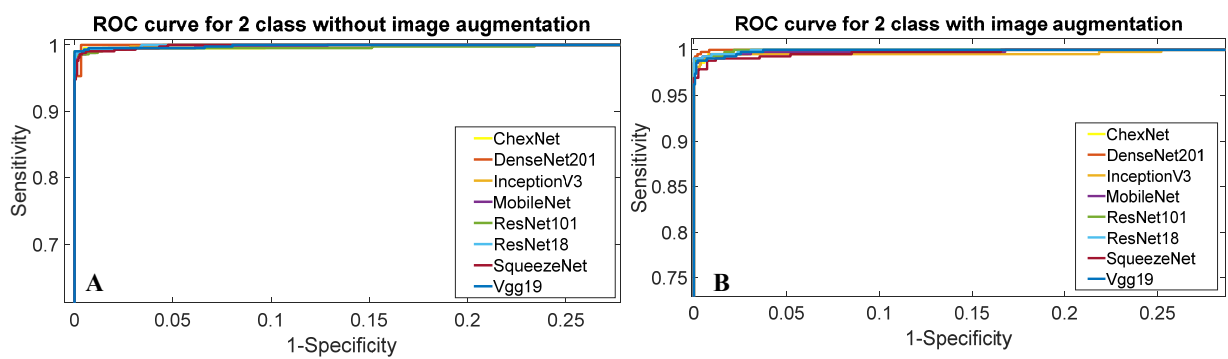

**Figure 5:** Comparison of the ROC curve for Normal, and COVID-19 Pneumonia classification using CNN based models without (A) and with (B) image augmentation.

X-ray image database and the network supposed to perform better for X-ray image classification without the requirement of training again on a larger dataset. However, in this classification problem as the COVID-19 images are significantly different from normal images all the tested networks are performing well. This is apparent from the ROC curves of Figure 5 as well. In both the cases (without and with augmentation) for two-class problem, ROC curves are showing comparable performance from all the networks.

*B. EXPERIMENTAL RESULTS – THREE CLASS PROBLEM*
Table 3 summarizes the performance matrix for different pre-trained CNN algorithms tested for the two different classification schemes without and with image augmentation. It can be noticed that all the pre-trained networks (shallow or deep) are showing very similar performance apart from CheXNet in case of training without image augmentation. If the pre-trained networks are trained on a small image dataset as reported by the most of the research groups in the literature, the performance difference is very marginal and overall performance is reduced for three-class problem in comparison to two-class problem. This is expected as networks are now confused between COVID-19 and viral pneumonia. However, CheXNet is still performing well while trained on a small dataset as CheXNet was originally trained one a very large X-ray image dataset. On the other hand, while the image augmentation was applied to the training image set, all the pre-trained networks are now performing based on their capability to distinguish the three-class images. Typically, the deeper the network the better is the performance in distinguishing the image classes.

TABLE 3
WEIGHTED AVERAGE PERFORMANCE METRICS FOR DIFFERENT DEEP LEARNING NETWORKS FOR THREE-CLASS CLASSIFICATION PROBLEM WITH AND WITHOUT IMAGE AUGMENTATION

| Schemes | Models | Accuracy | Precision (PPV) | Sensitivity (Recall) | F1 Scores | Specificity |
|---|---|---|---|---|---|---|
| Without image augmentation | SqueezeNet | 95.19 | 95.27 | 95.19 | 95.23 | 97.59 |
| | MobileNetv2 | 95.9 | 95.97 | 95.9 | 95.93 | 97.95 |
| | ResNet18 | 95.75 | 95.8 | 95.75 | 95.78 | 97.88 |
| | InceptionV3 | 94.96 | 94.98 | 94.95 | 94.96 | 97.49 |
| | ResNet101 | 95.36 | 95.4 | 95.36 | 95.38 | 97.68 |
| | **CheXNet** | **97.74** | **96.61** | **96.61** | **96.61** | **98.31** |
| | DenseNet201 | 95.19 | 95.06 | 95.9 | 95.04 | 97.87 |
| | VGG19 | 95.04 | 95.06 | 95.03 | 95.04 | 97.51 |
| With image augmentation | SqueezeNet | 95.10 | 95.18 | 95.10 | 95.14 | 97.17 |
| | MobileNetv2 | 96.22 | 96.25 | 96.22 | 96.23 | 97.80 |
| | ResNet18 | 96.44 | 96.48 | 96.44 | 96.46 | 97.91 |
| | InceptionV3 | 96.20 | 97.00 | 96.40 | 96.60 | 97.50 |
| | ResNet101 | 96.22 | 96.24 | 96.22 | 96.23 | 97.80 |
| | CheXNet | 96.94 | 96.43 | 96.42 | 96.42 | 97.29 |
| | **DenseNet201** | **97.94** | **97.95** | **97.94** | **97.94** | **98.80** |
| | VGG19 | 96.00 | 96.50 | 96.25 | 96.38 | 97.52 |

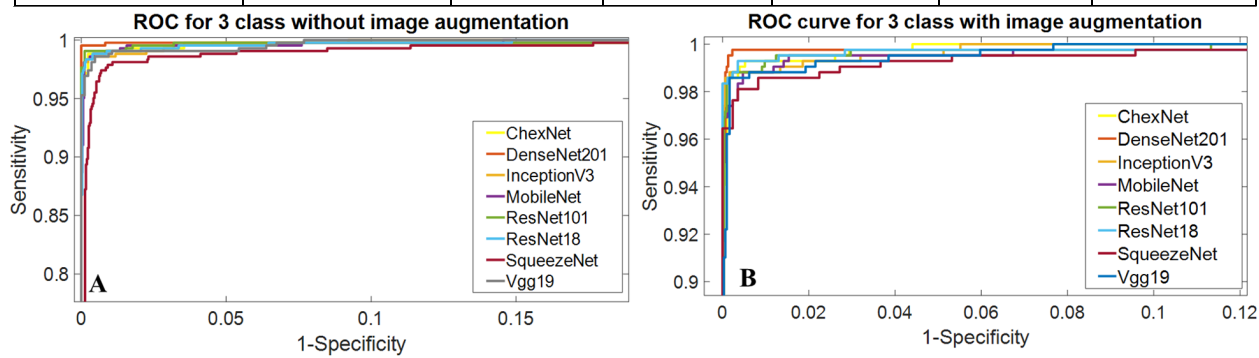

**Figure 6:** Comparison of the ROC curve for Normal, COVID-19 and viral Pneumonia classification using CNN based models without (A) and with (B) image augmentation.

However, it is important to note that Resnet18 and ResNet101 do not support this statement rather ResNet18 being a much shallower network than ResNet101, ResNet18 is still outperforming ResNet101.

Interestingly, CheXNet which is a 121-layer variant of DenseNet trained on X-ray images, is not outperforming a deeper variant of DenseNet with 201-layers. Therefore, it can be summarized that even though CheXNet was trained originally on X-ray images but training an even deeper network with a larger image set can give better chance of training from the new image sets on which the training is done, i.e., deep network can learn better and perform better if the training is carried out on a larger dataset. DenseNet201 outperforms other models in three-class classification scheme in terms of different performance indices when the image augmentation was employed and the performance matrix was significantly improved with image augmentation. It is obvious from Figure 6 that DenseNet201 with image augmentation can significantly increase overall network performance.

Figure 7 shows the confusion matrix for DenseNet201 for two-class and three-class problems with image augmentation. It is clear from Figure 7(A) that only three COVID-19 images out of 423 images were miss-classified to normal (false negative) and only three images out of 1579 images were miss-classified to COVID-19 (false positive). This reflects that this deep learning technique is extremely robust in distinguishing COVID-19 images from normal X-ray images. In the three-class problem, only one COVID-19 image was miss-classified to normal, which is one of the three images miss-classified by the two-class classifier. Two other COVID-19 images were miss-classified as viral pneumonia images. None of normal images were miss-classified to COVID-19 by the three-class classifier although several normal images were miss-classified to viral pneumonia. COVID-19 image miss-classified to normal has bad consequences than miss-classified to other disease category (i.e., viral pneumonia). Similarly, normal images miss-classified to viral pneumonia has less severe consequence than to be miss-classified to COVID-19 pneumonia. Only four viral pneumonia images were miss-classified to COVID-19 out of 1485 images while 33 images miss-classified to normal. It can be noted that network is not confusing between COVID-19, and other two image classes rather network is more confused between viral pneumonia and normal images. However, the high precision and F1 score show that the network is still performing excellent in classifying most of the images reliably. This is very important, as the computer-aided system (CAD) should not classify any COVID-19 patients to normal or vice versa; however, it is important to see the reason of the classifier being failed for three COVID-19 patients' X-ray images and miss-classified them to normal.

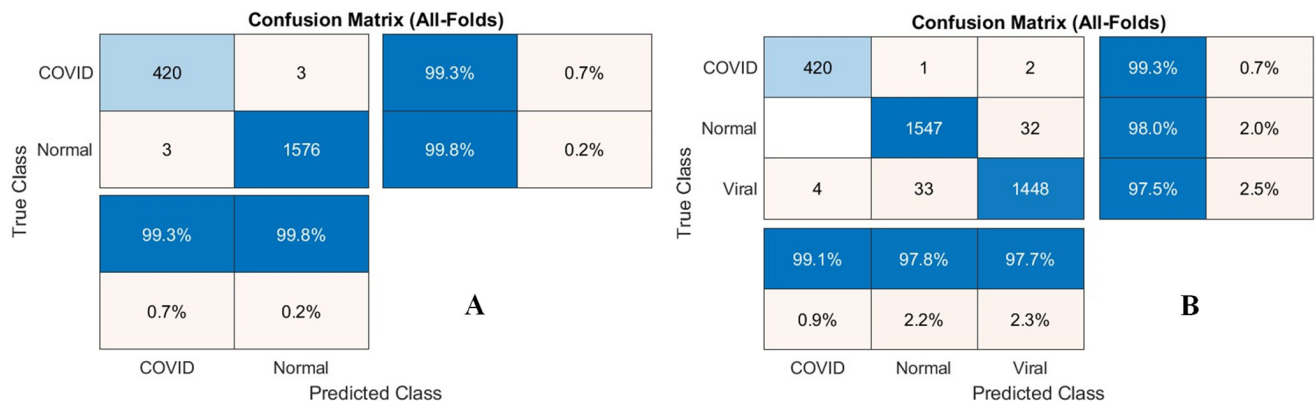

**Figure 7:** Confusion matrix for classification of (A) Normal and COVID-19, and (B) Normal, COVID-19 and Viral Pneumonia using DenseNet201.

The difference between normal and COVID-19 X-ray images can be observed in the deep convolutional layer of pre-trained CNN model. It is notable from Figure 8 that the 14th layer of the DenseNet201 can detect features that can distinguish normal, COVID-19 and Viral Pneumonia images. This shows the reason of the success of the network in detecting COVID-19 X-ray images and distinguishing it from normal and viral pneumonia images, which several groups of researchers reported earlier are not reliably possible by plain X-ray images [70-73]. It is really difficult for the practicing radiologist to find abnormality in the early stage of COVID-19. However, with the help of artificial intelligence, the X-ray images can be used to identify the deep layer features which are not visible to the human eyes[74]. The deep layers enhance the distinctive features of COVID-19, viral pneumonia and normal patients' X-ray images, thereby enhancing the chance of identifying the abnormality in the lungs of the patients.

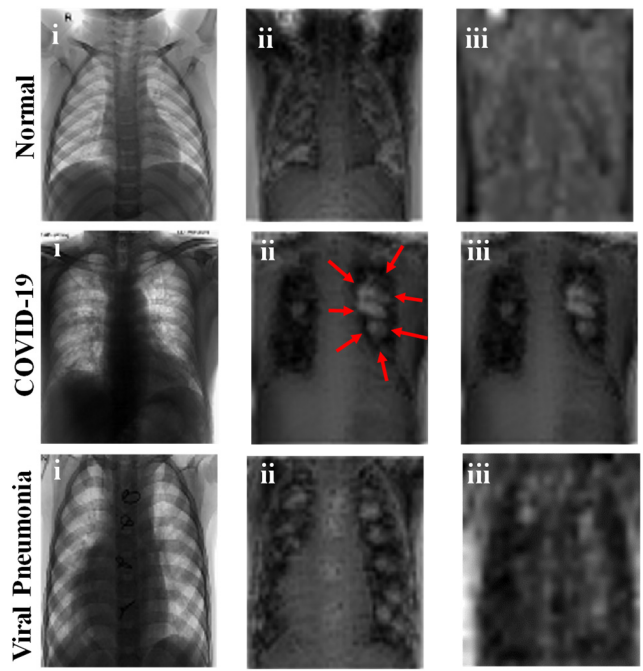

**Figure 8:** Images of 23rd channel of first convolutional layer (i), 14th convolutional layer (ii) and 29th convolutional layer images (iii) from DenseNet201 for different subject groups: Normal, COVID-19, and Viral Pneumonia. Red arrows in COVID-19 image shows the regions of light focus edge, a distinctive feature in COVID-19 patients' X-ray images which are not present in Viral Pneumonia and normal patients.

Figure 9 shows the three images of COVID-19 miss-classified to normal. Image 01-03 are miss-classified by two-class classifier and Image-03 is miss-classified by three-class classifier. The main reason behind the missing of these COVID-19 images is a less opacity in the left and right upper lobe and suprahilar on posterior-to-anterior x-ray images, which is very similar to normal X-ray images (see Figure 8). The algorithm fails if no evident light focus edge feature is appeared in the deep layer and this type of COVID-19 cases have to be confirmed by other techniques. These three images were evaluated by three practicing radiologists to identify what is their evaluations for these three images. First and third images were identified as no sign or very little sign of COVID-19 by the radiologists while image-02 was identified as very mild stage of lung infections. It can be summarized that the proposed technique can classify most of the COVID-19 X-ray images very reliably.

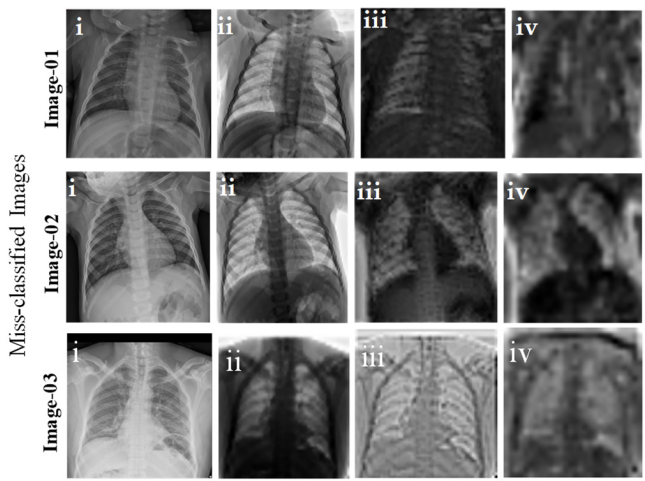

**Figure 9:** Three COVID-19 X-ray images which are miss-classified to normal images by two- and three-class classifier. Note: Image 01-03 are miss-classified by two-class classifier and Image-03 is miss-classified by three-class classifier.

## IV. CONCLUSION

This work presents deep CNN based transfer learning approach for automatic detection of COVID-19 pneumonia. Eight different popular and previously reported efficient CNN based deep learning algorithms were trained, validated and tested for classifying normal and pneumonia patients using chest X-ray images. It was observed that DenseNet201 outperforms other different deep CNN networks while image augmentation was used for training the CNN models. CheXNet which is a variant of DenseNet was outperforming other networks while image augmentation was not used. This is obvious as the CheXNet was pre-trained on a large X-ray database and it is showing better performance on this study while trained on a small non-augmented image dataset. However, a deeper version of DenseNet, when trained on a large augmented dataset, Dense201 outperforms CheXNet. This clearly reveals the fact that the performance reported on smaller database in the literature should be evaluated on a large dataset otherwise, the findings of these studies cannot be generalized for real applications. In this work, authors have reported the findings from a large database along with the image augmentation to train shallow and deep networks and it was observed that deep networks perform better than the shallow networks particularly in classifying normal and viral images as most of the networks can identify COVID-19 with very high sensitivity. The classification accuracy, precision, sensitivity, and specificity of normal and COVID-19 images, and normal, COVID-19 and viral pneumonia were (99.7%, 99.7%, 99.7% and 99.55%), and (97.9%, 97.95%, 97.9%, and 98.8%) respectively. COVID-19 has already become a threat to the world's healthcare system and economy and thousands of people have already died. Deaths were initiated by respiratory failure, which leads to the failure of other organs. Since a large number of patients attending out-door or emergency, doctor's time is limited and computer-aided-diagnosis can save lives by early screening and proper-care. Moreover, there is a large degree of variability in the input images from the X-ray machines due to the variations of expertise of the radiologist. Artificial intelligence exhibits an excellent performance in classifying COVID-19 pneumonia provided that the network is effectively trained from a large dataset. We believe that this computer aided diagnostic tool can significantly improve the speed and accuracy in the screening of COVID-19 positive cases. The method would be highly useful in this pandemic where disease burden and need for preventive measures are at odds with available resources.

## AUTHORS CONTRIBUTION

Muhammad E. H. Chowdhury: Conceptualization, Writing - Review & Editing, Supervision, and Project administration. Tawsifur Rahman: Data Curation, Methodology, Software, Validation, Formal analysis, Writing - Review & Editing. Amith Khandakar: Data Curation, Investigation, Resources, Writing - Original Draft, Writing - Review & Editing. Rashid Mazhar: Writing - Original Draft, Writing - Review & Editing. Muhammad Abdul Kadir: Methodology, Visualization, Editing. Zaid Bin Mahbub: Methodology, Visualization. Khandakar R. Islam: Data Curation, Writing - Original Draft. Muhammad Salman Khan: Visualization, Writing - Original Draft. Atif Iqbal: Writing - Review & Editing, Nasser Al-Emadi: Writing - Review & Editing, Supervision, Mamun Bin Ibne Reaz: Writing - Review & Editing, Supervision, Conceptualization. M. T. Islam: Writing - Review & Editing, Supervision.

## FUNDING

The publication of this article was funded by the Qatar National Library and this work was made possible by NPRP12S-0227-190164 from the Qatar National Research Fund, a member of Qatar Foundation, Doha, Qatar. The statements made herein are solely the responsibility of the authors.

## ACKNOWLEDGMENTS

The authors would like to thank Italian Society of Medical Radiology and Interventional for sharing the X-ray images of COVID-19 patients publicly and would like to thank J. P. Cohen for taking the initiative to gather images from articles and online resources. Last but not the least, authors would like to acknowledge the Chest X-Ray Images (pneumonia) database and RSNA Pneumonia Detection Challenge in Kaggle which helped significantly to make this work possible. Otherwise, normal and viral pneumonia images were not accessible to the team.

## CONFLICTS OF INTEREST
The authors declare no conflict of interest.

## REFERENCE
[1] (2020). WHO Director-General's opening remarks at the media briefing on COVID-19 - 11 March 2020. Available: https://www.who.int/dg/speeches/detail/who-director-general-s-opening-remarks-at-the-media-briefing-on-covid-19---11-march-2020
[2] (2020). Coronavirus Disease 2019 (COVID-19). Available: https://www.cdc.gov/coronavirus/2019-ncov/need-extra-precautions/people-at-higher-risk.html
[3] W. H. Organization, "Global COVID-19 report," March 25,2020 2020.
[4] J. H. U. MEDICINE. (2020). Coronavirus COVID-19 Global Cases by the Center for Systems Science and Engineering (CSSE) at Johns Hopkins University (JHU). Available: https://coronavirus.jhu.edu/map.html
[5] W. Wang, Y. Xu, R. Gao, R. Lu, K. Han, G. Wu, et al., "Detection of SARS-CoV-2 in Different Types of Clinical Specimens," Jama, 2020.
[6] T. Yang, Y.-C. Wang, C.-F. Shen, and C.-M. Cheng, "Point-of-Care RNA-Based Diagnostic Device for COVID-19," ed: Multidisciplinary Digital Publishing Institute, 2020.
[7] A. J. NEWS. (2020). India's poor testing rate may have masked coronavirus cases. Available: https://www.aljazeera.com/news/2020/03/india-poor-testing-rate-masked-coronavirus-cases-200318040314568.html
[8] A. J. NEWS. (2020). Bangladesh scientists create $3 kit. Can it help detect COVID-19? Available: https://www.aljazeera.com/news/2020/03/bangladesh-scientists-create-3-kit-detect-covid-19-200323035631025.html
[9] N. Wetsman. (2020). CORONAVIRUS TESTING SHOULDN'T BE THIS COMPLICATED. Available: https://www.theverge.com/2020/3/17/21184015/coronavirus-testing-pcr-diagnostic-point-of-care-cdc-techonology
[10] D. Wang, B. Hu, C. Hu, F. Zhu, X. Liu, J. Zhang, et al., "Clinical characteristics of 138 hospitalized patients with 2019 novel coronavirus–infected pneumonia in Wuhan, China," Jama, 2020.
[11] N. Chen, M. Zhou, X. Dong, J. Qu, F. Gong, Y. Han, et al., "Epidemiological and clinical characteristics of 99 cases of 2019 novel coronavirus pneumonia in Wuhan, China: a descriptive study," The Lancet, vol. 395, pp. 507-513, 2020.
[12] Q. Li, X. Guan, P. Wu, X. Wang, L. Zhou, Y. Tong, et al., "Early transmission dynamics in Wuhan, China, of novel coronavirus–infected pneumonia," New England Journal of Medicine, v. 382, pp.1199-1207 2020.
[13] C. Huang, Y. Wang, X. Li, L. Ren, J. Zhao, Y. Hu, et al., "Clinical features of patients infected with 2019 novel coronavirus in Wuhan, China," The Lancet, vol. 395, pp. 497-506, 2020.
[14] V. M. Corman, O. Landt, M. Kaiser, R. Molenkamp, A. Meijer, D. K. Chu, et al., "Detection of 2019 novel coronavirus (2019-nCoV) by real-time RT-PCR," Eurosurveillance, vol. 25, 2020.
[15] D. K. Chu, Y. Pan, S. Cheng, K. P. Hui, P. Krishnan, Y. Liu, et al., "Molecular diagnosis of a novel coronavirus (2019-nCoV) causing an outbreak of pneumonia," Clinical chemistry, vol. 66(4), pp.549-555, 2020.
[16] N. Zhang, L. Wang, X. Deng, R. Liang, M. Su, C. He, et al., "Recent advances in the detection of respiratory virus infection in humans," Journal of medical virology, vol. 92, pp. 408-417, 2020.
[17] M. Chung, A. Bernheim, X. Mei, N. Zhang, M. Huang, X. Zeng, et al., "CT imaging features of 2019 novel coronavirus (2019-nCoV)," Radiology, vol. 295, pp.202–207, 2020.
[18] M. Hosseiny, S. Kooraki, A. Gholamrezanezhad, S. Reddy, and L. Myers, "Radiology perspective of coronavirus disease 2019 (COVID-19): lessons from severe acute respiratory syndrome and Middle East respiratory syndrome," American Journal of Roentgenology, vol. 214(5), pp.1078-1082, 2020.
[19] S. Salehi, A. Abedi, S. Balakrishnan, and A. Gholamrezanezhad, "Coronavirus Disease 2019 (COVID-19): A Systematic Review of Imaging Findings in 919 Patients," American Journal of Roentgenology, vol. 215, pp. 1-7, 2020.
[20] A. M. Tahir, M. E. Chowdhury, A. Khandakar, S. Al-Hamouz, M. Abdalla, S. Awadallah, et al., "A systematic approach to the design and characterization of a smart insole for detecting vertical ground reaction force (vGRF) in gait analysis," Sensors, vol. 20, p. 957, 2020.
[21] M. E. Chowdhury, K. Alzoubi, A. Khandakar, R. Khallifa, R. Abouhasera, S. Koubaa, et al., "Wearable real-time heart attack detection and warning system to reduce road accidents," Sensors, vol. 19, p. 2780, 2019.
[22] M. E. Chowdhury, A. Khandakar, K. Alzoubi, S. Mansoor, A. M Tahir, M. B. I. Reaz, et al., "Real-Time Smart-Digital Stethoscope System for Heart Diseases Monitoring," Sensors, vol. 19, p. 2781, 2019.
[23] K. Kallianos, J. Mongan, S. Antani, T. Henry, A. Taylor, J. Abuya, et al., "How far have we come? Artificial intelligence for chest radiograph interpretation," Clinical radiology, vol. 74(5), pp.338-345, 2019.
[24] M. Dahmani, M. E. Chowdhury, A. Khandakar, T. Rahman, K. Al-Jayyousi, A. Hefny, et al., "An Intelligent and Low-cost Eye-tracking System for Motorized Wheelchair Control," arXiv preprint arXiv:2005.02118, 2020.
[25] T. Rahman, M. E. Chowdhury, A. Khandakar, K. R. Islam, K. F. Islam, Z. B. Mahbub, et al., "Transfer Learning with Deep Convolutional Neural Network (CNN) for Pneumonia Detection using Chest X-ray," Applied Sciences, vol. 10, p. 3233, 2020.
[26] A. Krizhevsky, I. Sutskever, and G. E. Hinton, "Imagenet classification with deep convolutional neural networks," in Advances in neural information processing systems, 2012, pp. 1097-1105.
[27] P. Gómez, M. Semmler, A. Schützenberger, C. Bohr, and M. Döllinger, "Low-light image enhancement of high-speed endoscopic videos using a convolutional neural network," Medical & biological engineering & computing, vol. 57, pp. 1451-1463, 2019.
[28] J. Choe, S. M. Lee, K.-H. Do, G. Lee, J.-G. Lee, S. M. Lee, et al., "Deep Learning–based Image Conversion of CT Reconstruction Kernels Improves Radiomics Reproducibility for Pulmonary Nodules or Masses," Radiology, vol. 292, pp. 365-373, 2019.
[29] D. S. Kermany, M. Goldbaum, W. Cai, C. C. Valentim, H. Liang, S. L. Baxter, et al., "Identifying medical diagnoses and treatable diseases by image-based deep learning," Cell, vol. 172, pp. 1122-1131. e9, 2018.
[30] M. Negassi, R. Suarez-Ibarrola, S. Hein, A. Miernik, and A. Reiterer, "Application of artificial neural networks for automated analysis of cystoscopic images: a review of the current status and future prospects," World Journal of Urology, pp. 1-10, 2020.
[31] P. Wang, X. Xiao, J. R. G. Brown, T. M. Berzin, M. Tu, F. Xiong, et al., "Development and validation of a deep-learning algorithm for the detection of polyps during colonoscopy," Nature biomedical engineering, vol. 2, pp. 741-748, 2018.
[32] V. Chouhan, S. K. Singh, A. Khamparia, D. Gupta, P. Tiwari, C. Moreira, et al., "A Novel Transfer Learning Based Approach for Pneumonia Detection in Chest X-ray Images," Applied Sciences, vol. 10, p. 559, 2020.
[33] D. Gershgorn. (2017). The data that transformed AI research—and possibly the world. Available: https://qz.com/1034972/the-data-that-changed-the-direction-of-ai-research-and-possibly-the-world/
[34] X. Gu, L. Pan, H. Liang, and R. Yang, "Classification of bacterial and viral childhood pneumonia using deep learning in chest radiography," in Proceedings of the 3rd International Conference on Multimedia and Image Processing, 2018, pp. 88-93.
[35] X. Wang, Y. Peng, L. Lu, Z. Lu, M. Bagheri, and R. Summers, "Hospital-scale Chest X-ray Database and Benchmarks on Weakly-Supervised Classification and Localization of Common Thorax Diseases," in IEEE CVPR, 2017.
[36] O. Ronneberger, P. Fischer, and T.-n. Brox, "Convolutional networks for biomedical image segmentation," in Paper presented at: International Conference on Medical Image Computing and Computer-Assisted Intervention, 2015.
[37] P. Rajpurkar, J. Irvin, R. L. Ball, K. Zhu, B. Yang, H. Mehta, et al., "Deep learning for chest radiograph diagnosis: A retrospective comparison of the CheXNeXt algorithm to practicing radiologists," PLoS medicine, vol. 15, p. e1002686, 2018.
[38] T. K. K. Ho and J. Gwak, "Multiple feature integration for classification of thoracic disease in chest radiography," Applied Sciences, vol. 9, p. 4130, 2019.

[39] P. Lakhani and B. Sundaram, "Deep learning at chest radiography: automated classification of pulmonary tuberculosis by using convolutional neural networks," Radiology, vol. 284, pp. 574-582, 2017.

[40] A. w. Linda wang, "COVID-Net: A Tailored Deep Convolutional Neural Network Design for Detection of COVID-19 Cases from Chest Radiography Images," arXiv preprint arXiv:2003.09871, 2020.

[41] S. Wang, B. Kang, J. Ma, X. Zeng, M. Xiao, J. Guo, et al., "A deep learning algorithm using CT images to screen for Corona Virus Disease (COVID-19)," medRxiv, 2020.

[42] A. S. Joaquin. (2020). Using Deep Learning to detect Pneumonia caused by NCOV-19 from X-Ray Images. Available: https://towardsdatascience.com/using-deep-learning-to-detect-ncov-19-from-x-ray-images-1a89701d1acd

[43] I. D. Apostolopoulos and T. A. Mpesiana, "Covid-19: automatic detection from x-ray images utilizing transfer learning with convolutional neural networks," Physical and Engineering Sciences in Medicine, vol. 43, pp.635–640, 2020.

[44] A. Abbas, M. M. Abdelsamea, and M. M. Gaber, "Classification of COVID-19 in chest X-ray images using DeTraC deep convolutional neural network," arXiv preprint arXiv:2003.13815, 2020.

[45] E. E.-D. Hemdan, M. A. Shouman, and M. E. Karar, "Covidx-net: A framework of deep learning classifiers to diagnose covid-19 in x-ray images," arXiv preprint arXiv:2003.11055, 2020.

[46] A. Narin, C. Kaya, and Z. Pamuk, "Automatic detection of coronavirus disease (covid-19) using x-ray images and deep convolutional neural networks," arXiv preprint arXiv:2003.10849, 2020.

[47] J. Zhang, Y. Xie, Y. Li, C. Shen, and Y. Xia, "Covid-19 screening on chest x-ray images using deep learning based anomaly detection," arXiv preprint arXiv:2003.12338, 2020.

[48] P. K. Sethy and S. K. Behera, "Detection of coronavirus disease (covid-19) based on deep features," Preprints, vol. 2020030300, p. 2020, 2020.

[49] P. Afshar, S. Heidarian, F. Naderkhani, A. Oikonomou, K. N. Plataniotis, and A. Mohammadi, "Covid-caps: A capsule network-based framework for identification of covid-19 cases from x-ray images," arXiv preprint arXiv:2004.02696, 2020.

[50] T. Ozturk, M. Talo, E. A. Yildirim, U. B. Baloglu, O. Yildirim, and U. R. Acharya, "Automated detection of COVID-19 cases using deep neural networks with X-ray images," Computers in Biology and Medicine, vol. 121, p. 103792, 2020.

[51] L. Wang and A. Wong, "COVID-Net: A Tailored Deep Convolutional Neural Network Design for Detection of COVID-19 Cases from Chest X-Ray Images," arXiv preprint arXiv:2003.09871, 2020.

[52] F. Ucar and D. Korkmaz, "COVIDiagnosis-Net: Deep Bayes-SqueezeNet based Diagnostic of the Coronavirus Disease 2019 (COVID-19) from X-Ray Images," Medical Hypotheses, vol. 140, p. 109761, 2020.

[53] L. O. Hall, R. Paul, D. B. Goldgof, and G. M. Goldgof, "Finding covid-19 from chest x-rays using deep learning on a small dataset," arXiv preprint arXiv:2004.02060, 2020.

[54] H. S. Maghdid, A. T. Asaad, K. Z. Ghafoor, A. S. Sadiq, and M. K. Khan, "Diagnosing COVID-19 pneumonia from X-ray and CT images using deep learning and transfer learning algorithms," arXiv preprint arXiv:2004.00038, 2020.

[55] S. Minaee, R. Kafieh, M. Sonka, S. Yazdani, and G. J. Soufi, "Deep-covid: Predicting covid-19 from chest x-ray images using deep transfer learning," arXiv preprint arXiv:2004.09363, 2020.

[56] X. Li and D. Zhu, "Covid-xpert: An ai powered population screening of covid-19 cases using chest radiography images," arXiv preprint arXiv:2004.03042, 2020.

[57] A. I. Khan, J. L. Shah, and M. M. Bhat, "Coronet: A deep neural network for detection and diagnosis of COVID-19 from chest x-ray images," Computer Methods and Programs in Biomedicine, vol.186, p. 105581, 2020.

[58] J. Deng, W. Dong, R. Socher, L.-J. Li, K. Li, and L. Fei-Fei, "Imagenet: A large-scale hierarchical image database," in 2009 IEEE conference on computer vision and pattern recognition, 2009, pp. 248-255.

[59] N. Tajbakhsh, J. Y. Shin, S. R. Gurudu, R. T. Hurst, C. B. Kendall, M. B. Gotway, et al., "Convolutional neural networks for medical image analysis: Full training or fine tuning?," IEEE transactions on medical imaging, vol. 35, pp. 1299-1312, 2016.

[60] S. J. Pan and Q. Yang, "A survey on transfer learning," IEEE Transactions on knowledge and data engineering, vol. 22, pp. 1345-1359, 2009.

[61] T. R. Muhammad E. H. Chowdhury, Amith Khandakar, Rashid Mazhar, Muhammad Abdul Kadir, Zaid Bin Mahbub, Khandakar R. Islam, Muhammad Salman Khan, Atif Iqbal, Nasser Al-Emadi, Mamun Bin Ibne Reaz. (2020). COVID-19 CHEST X-RAY DATABASE. Available: https://www.kaggle.com/tawsifurrahman/covid19-radiography-database

[62] S.-I. S. o. M. a. I. Radiology. (2020). COVID-19 Database. Available: https://www.sirm.org/category/senza-categoria/covid-19/

[63] J. C. Monteral. (2020). COVID-Chestxray Database. Available: https://github.com/ieee8023/covid-chestxray-dataset

[64] (2020). Radiopedia, Available: https://radiopaedia.org/search?lang=us&page=4&q=covid+19&scope=all&utf8=%E2%9C%93

[65] C. Imaging. (2020). This is a thread of COVID-19 CXR (all SARS-CoV-2 PCR+) from my hospital (Spain). I hope it could help. Available: https://threadreaderapp.com/thread/1243928581983670272.html

[66] X. Wang, Y. Peng, L. Lu, Z. Lu, M. Bagheri, and R. M. Summers, "Chestx-ray8: Hospital-scale chest x-ray database and benchmarks on weakly-supervised classification and localization of common thorax diseases," in Proceedings of the IEEE conference on computer vision and pattern recognition, 2017, pp. 2097-2106.

[67] P. Mooney. (2018). Chest X-Ray Images (Pneumonia). Available: https://www.kaggle.com/paultimothymooney/chest-xray-pneumonia

[68] Y. LeCun, K. Kavukcuoglu, and C. Farabet, "Convolutional networks and applications in vision," in Proceedings of 2010 IEEE international symposium on circuits and systems, 2010, pp. 253-256.

[69] ResNet, AlexNet, VGGNet, Inception: Understanding various architectures of Convolutional Networks. Available: https://cv-tricks.com/cnn/understand-resnet-alexnet-vgg-inception/

[70] W. Xia, J. Shao, Y. Guo, X. Peng, Z. Li, and D. Hu, "Clinical and CT features in pediatric patients with COVID-19 infection: Different points from adults," Pediatric pulmonology, vol. 55, pp. 1169-1174, 2020.

[71] A. Filatov, P. Sharma, F. Hindi, and P. S. Espinosa, "Neurological complications of coronavirus disease (COVID-19): encephalopathy," Cureus, vol. 12, 2020.

[72] J. Lim, S. Jeon, H.-Y. Shin, M. J. Kim, Y. M. Seong, W. J. Lee, et al., "Case of the index patient who caused tertiary transmission of coronavirus disease 2019 in Korea: The application of lopinavir/ritonavir for the treatment of COVID-19 pneumonia monitored by quantitative RT-PCR," Journal of Korean Medical Science, vol. 35, p. e79, 2020.

[73] M. B. Weinstock, A. Echenique, J. W. R. DABR, A. Leib, and F. A. ILLUZZI, "Chest x-ray findings in 636 ambulatory patients with COVID-19 presenting to an urgent care center: a normal chest x-ray is no guarantee," J Urgent Care Med, vol. 14, pp. 13-8, 2020.

[74] A. D. Mete Ahishali, Mehmet Yamac, Serkan Kiranyaz, Muhammad E. H. Chowdhury, Khalid Hameed, Tahir Hamid, Rashid Mazhar and Moncef Gabbouj, "A Comparative Study on Early Detection of COVID-19 from Chest X-Ray Images," arXiv:2006.05332[cs],Jun. 2020.